\documentclass[runningheads]{llncs}
\usepackage[T1]{fontenc}
\usepackage{graphicx}
\usepackage[numbers,sort&compress,sectionbib]{natbib}
\usepackage{url}
\usepackage[caption=false]{subfig}

\makeatletter \renewcommand\@biblabel[1]{#1.} \makeatother

\usepackage{tabularray}

\begin{document}

\title{Harmonizing Feature Attributions Across Deep Learning Architectures: Enhancing Interpretability and Consistency}
\titlerunning{Harmonizing Feature Attributions Across Deep Learning Architectures}
\author{Md Abdul Kadir\inst{1}\orcidID{0000-0002-8420-2536} \and
GowthamKrishna Addluri\inst{1}\orcidID{0009-0008-0513-9016} \and
Daniel Sonntag\inst{1, 2}\orcidID{0000-0002-8857-8709}}
\authorrunning{M. A. Kadir et al.}
\institute{German Research Center for Artificial Intelligence (DFKI), Germany\\
\email{\{abdul.kadir, Gowthamkrishna.Addluri, daniel.sonntag\}@dfki.de}\and
University of Oldenburg, Oldenburg, Germany}
\maketitle              %
\begin{abstract}
Enhancing the interpretability and consistency of machine learning models is critical to their deployment in real-world applications. Feature attribution methods have gained significant attention, which provide local explanations of model predictions by attributing importance to individual input features. This study examines the generalization of feature attributions across various deep learning architectures, such as convolutional neural networks (CNNs) and vision transformers. We aim to assess the feasibility of utilizing a feature attribution method as a future detector and examine how these features can be harmonized across multiple models employing distinct architectures but trained on the same data distribution. By exploring this harmonization, we aim to develop a more coherent and optimistic understanding of feature attributions, enhancing the consistency of local explanations across diverse deep-learning models. Our findings highlight the potential for harmonized feature attribution methods to improve interpretability and foster trust in machine learning applications, regardless of the underlying architecture.

\end{abstract}
\section{Introduction}
Deep learning models have revolutionized various domains, but their complex nature often hampers our ability to understand their decision-making processes \cite{alirezazadeh2022,kolesnikov20}. Interpretability techniques have emerged, with local and global explanations being two significant categories \cite{doshivelez2017rigorous}. Local explanations focus on understanding individual predictions, highlighting the most influential features for a specific instance. This method is valuable for understanding model behavior at a granular level and providing intuitive explanations for specific predictions. On the other hand, global explanations aim to capture overall model behavior and identify patterns and trends across the entire dataset. They offer a broader perspective and help uncover essential relationships between input features and model prediction. This paper delves into interpretability in deep learning models, particularly model-agnostic feature attribution, a subset of local explanation techniques.

Feature attribution refers to assigning importance or relevance to input features in a machine learning model's decision-making process \cite{ancona2018unified}. It aims to understand which features have the most significant influence on the model's predictions or outputs. Feature attribution techniques provide insights into the relationship between input features and the model's decision, shedding light on the factors that drive specific outcomes. These techniques are precious for interpreting complex models like deep learning, where the learned representations may be abstract and difficult to interpret directly \cite{samek2021}. By quantifying the contribution of individual features, feature attribution allows us to identify the most influential factors, validate the model's behavior, detect biases, and gain a deeper understanding of the decision-making process.

Feature attribution methods can be evaluated through various approaches and metrics \cite{samek2021}. Qualitative evaluation involves visually inspecting the attributions and assessing their alignment with domain knowledge. Perturbation analysis tests the sensitivity of attributions to changes in input features \cite{samek_evaluating_2017}. Sanity checks ensure the reasonableness of attributions, especially in classification problems. From a human perspective, we identify objects in images by recognizing distinct features \cite{mozer2001}. Similarly, deep learning models are trained to detect features from input data and make predictions based on these characteristics \cite{Schulz2012}. The primary objective of deep learning models, irrespective of the specific architecture, is to learn the underlying data distribution and capture unique identifying features for each class in the dataset.

Various deep learning architectures have proven proficient in capturing essential data characteristics within the training distribution \cite{tan_2019eff}. We assume that if a set of features demonstrates discriminative qualities for one architecture, it should likewise exhibit discriminative properties for a different architecture, provided both architectures are trained on the same data. This assumption forms the foundation for the consistency and transferability of feature attributions across various deep learning architectures.

Our experiments aim to explore the generalizability of features selected by a feature attribution method for one deep learning architecture compared to other architectures trained on the same data distribution. We refer to this process as harmonizing feature attributions across different architectures. Our experimental results also support our assumption and indicate that different architectures trained on the same data have a joint feature identification capability.

\section{Related work}
Various explanation algorithms have been developed better to understand the internal mechanisms of deep learning models. These algorithms, such as feature attribution maps, have gained significant popularity in deep learning research. They offer valuable insights into the rationale behind specific predictions made by deep learning models \cite{margaret_2020_xai}. Notable examples of these explanation methods include layer-wise relevance propagation \cite{montavon_methods_2018}, Grad-CAM \cite{selvaraju_2017_gradcam}, integrated gradient \cite{sundararajan_2017_ig}, guided back-propagation \cite{springenberg_2015_gb}, pixel-wise decomposition \cite{bach_pixel-wise_2015}, and contrastive explanations \cite{luss_leveraging_2021}.

Various methods have been developed to evaluate feature attribution maps. Ground truth data, such as object-localization or masks, has been used for evaluation \cite{chattopadhay_2018_gcampp,nunnari_2021_overlap}. Another approach focuses on the faithfulness of explanations, measuring how well they reflect the model's attention \cite{samek_evaluating_2017}. The IROF technique divides images into segments and evaluates explanations based on segment relevance \cite{rieger_irof_2020}. Pixel-wise evaluations involve flipping pixels or assessing attribution quality using pixel-based metrics \cite{bach_pixel-wise_2015,samek_evaluating_2017}.

\citet{chen_2021_feature} has demonstrated the utility of feature attribution methods for feature selection. Additionally, research conducted by \citet{morcos_2018} and \citet{kornblith_2019} has explored the internal representation similarity between different architectures. However, to the best of our knowledge, the generalization of feature attributions across diverse neural architectures still needs to be explored.

Motivated by the goal of evaluating feature attributions, we are investigating a novel approach that involves assessing feature attributions across multiple models belonging to different architectural designs. This method aims to provide a more comprehensive understanding of feature attribution in various contexts, thereby enhancing the overall explainability of deep neural networks.

\section{Methodology}
This experiment investigates the generalizability and transferability of feature attributions across different deep learning architectures trained on the same data distribution.
The experimental process involves generating feature attribution maps for a pretrained model, extracting features from input images, and passing them to two models with distinct architectures. The accuracy and output probability distribution are then calculated for each architecture.

\begin{figure}
\centering
\begin{tabular}{cccc}
\includegraphics[width=0.32\textwidth]{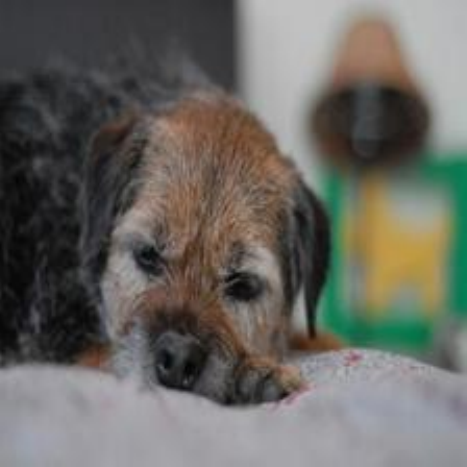} & \includegraphics[width=0.32\textwidth]{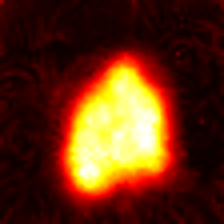} & \includegraphics[width=0.32\textwidth]{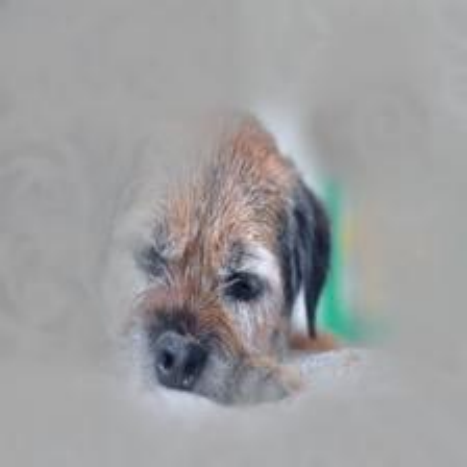} & \\
\includegraphics[width=0.32\textwidth ]{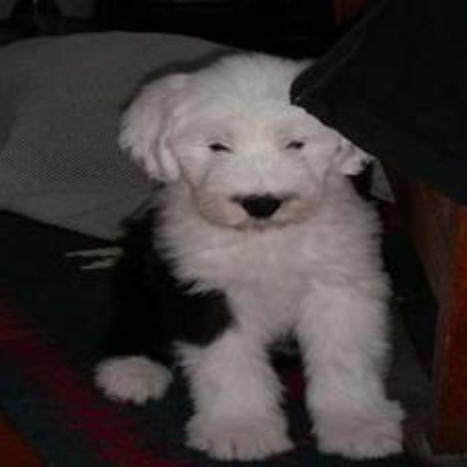} & \includegraphics[width=0.32\textwidth]{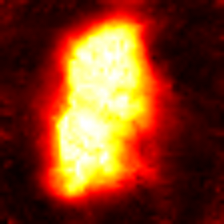} & \includegraphics[width=0.32\textwidth]{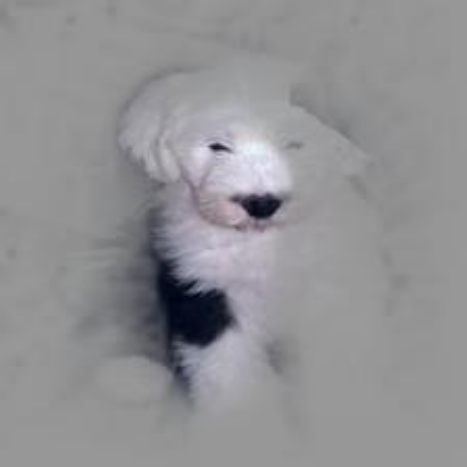} &\\
\includegraphics[width=0.32\textwidth]{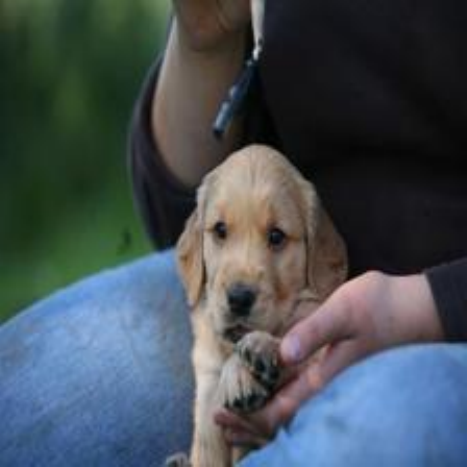} & \includegraphics[width=0.32\textwidth]{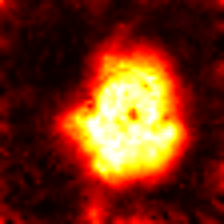} & \includegraphics[width=0.32\textwidth]{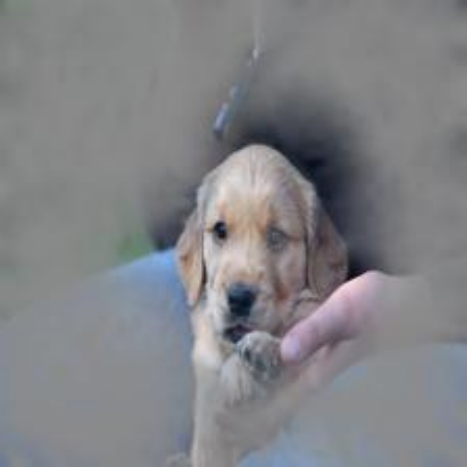} & \\
(a) Input image & (b) Feature attribution & (c) Features
\end{tabular}
\caption{Columns a, b, and c represent the input image, the feature attribution map generated by the soundness saliency algorithm, and the extracted features of the image based on the feature attribution map, respectively.}
\label{fig:image_grid}
\end{figure}

In this experiment, we employ a modified version of the Soundness Saliency (SS) method \cite{Gupta2022NewDA} for generating explanations.
The primary objective with a network $f$, for a specific input $x$ (Fig. \ref{fig:image_grid} (a)), and label $a$, is to acquire a map or mask $M \in \{0, 1\}^{hw}$.

This map aims to minimize the expectation $E_{\tilde{x} \sim {(x, M)}}  [- \sum f_{i}(\tilde{x})log(f_{i}(\tilde{x}))]$, wherein the probability assigned by the network to a modified or composite input $x$ is maximized.

\begin{equation}
    \tilde{x} \sim \Gamma(x, a) \equiv \overline{x} \sim \mathcal{X}, \tilde{x} = M \odot x + (1- M) \odot \overline{x}
\end{equation}
Here, $M$ (Fig. \ref{fig:image_grid} (b)) represents the feature attribution map generated by the Soundness Saliency algorithm. The saliency map $M$ provides information about the importance of each pixel and the extent of its contribution to the classification. If the value of $M$ (Fig. \ref{fig:image_grid} (b)) for a specific pixel is 0, it implies that the pixel has no significance in the classification process. Conversely, if the value of $M$ for a particular pixel is high, it indicates that the pixel is highly important for the classification. We enhance the extraction of important features (Fig \ref{fig:image_grid} (c)) from input by applying the Hadamard product between each input channel and the corresponding attribution map $M$.
In addition to the Soundness Saliency (SS) algorithm, we also employ Grad-CAM \cite{selvaraju_2017_gradcam} (GC) for feature extraction.
We utilize the selected features Fig. \ref{fig:image_grid} (c) extracted through feature attributions and feed them to two distinct models with different architectures, albeit trained on the same training data. Accuracy, F1 score, and output probability scores are calculated for these models. The focus is observing model prediction changes when only the selected features are inputted rather than the entire image.

\section{Experiment and Results}
In this study, we selected four distinct pretrained architectures: the Vision Transformer architecture (ViT) \cite{kolesnikov20}, EfficientNet-B7 (E-7) \cite{tan_2019eff}, EfficientNet-B6 (E-6)  \cite{tan_2019eff}, and EfficientNet-B5 (E-5)  \cite{tan_2019eff}. To generate feature attribution maps, we first employed E-7 along with a challenging subset\footnote{\url{https://github.com/fastai/imagenette}} of the ImageNet validation data, which is known to be particularly difficult for classifiers. 

Subsequently, we generated feature maps for all test data and passed them to E-6 and E-5. In parallel, we also generated features for ViT and followed the same procedure. We chose to utilize both a transformer and a CNN architecture in our experiments because they are fundamentally different from one another, allowing for a comprehensive evaluation of the various architectures.

\begin{figure}
    \centering
    \begin{tabular}{cccccc}
        \subfloat[E-7 (I)]{\includegraphics[width=0.16\textwidth]{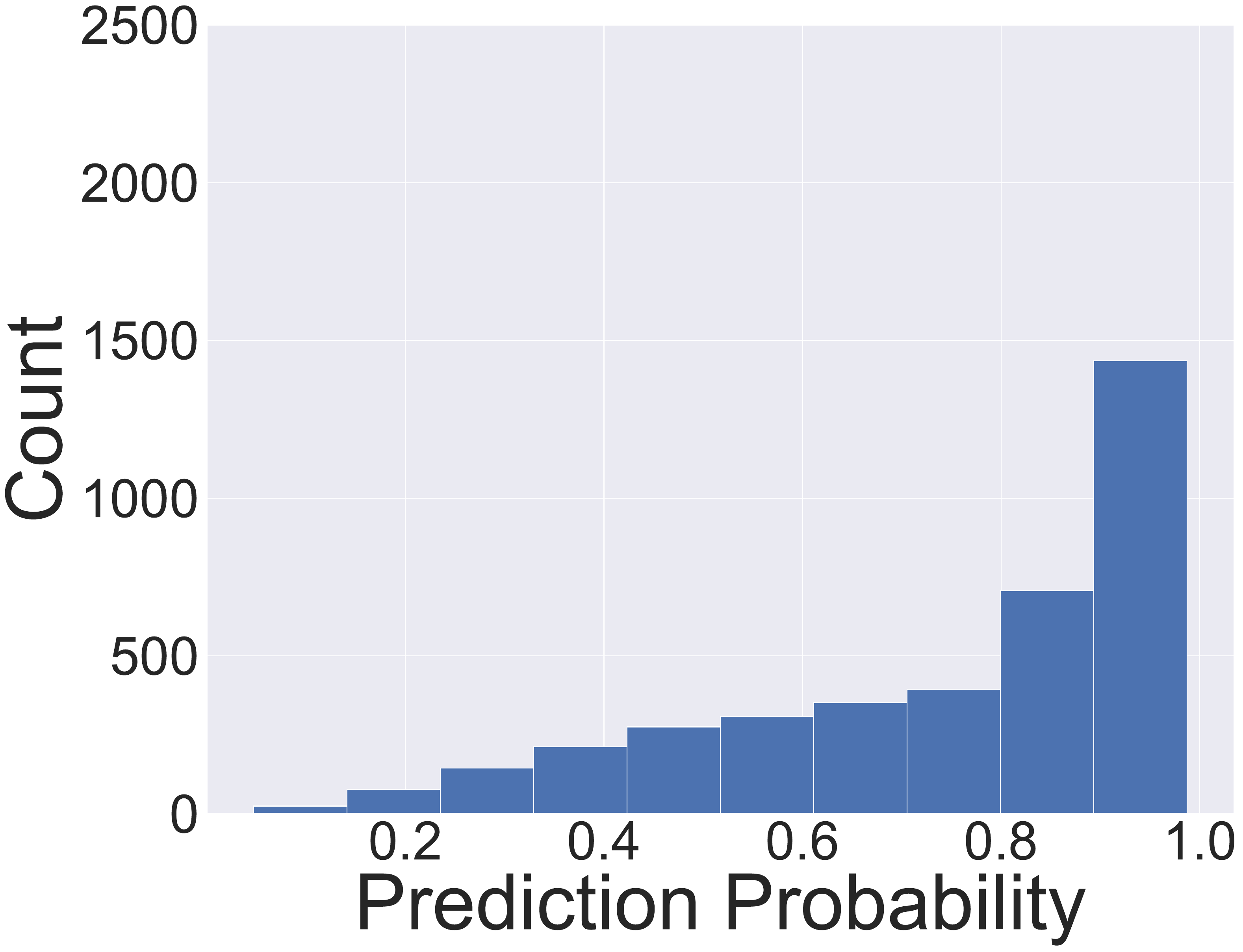}} &
        \subfloat[E-7 (F)]{\includegraphics[width=0.16\textwidth]{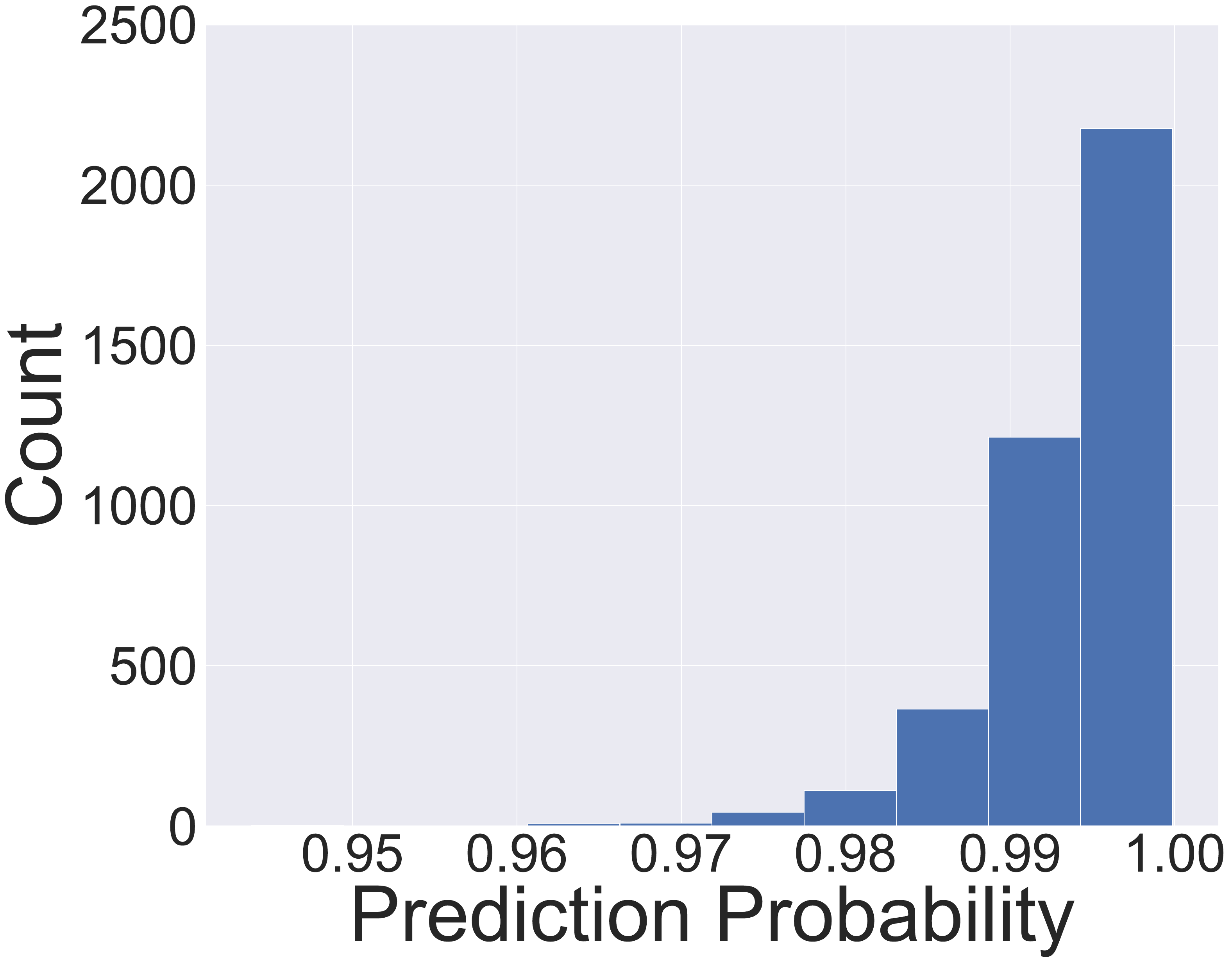}} &
        \subfloat[E-6 (I)]{\includegraphics[width=0.16\textwidth]{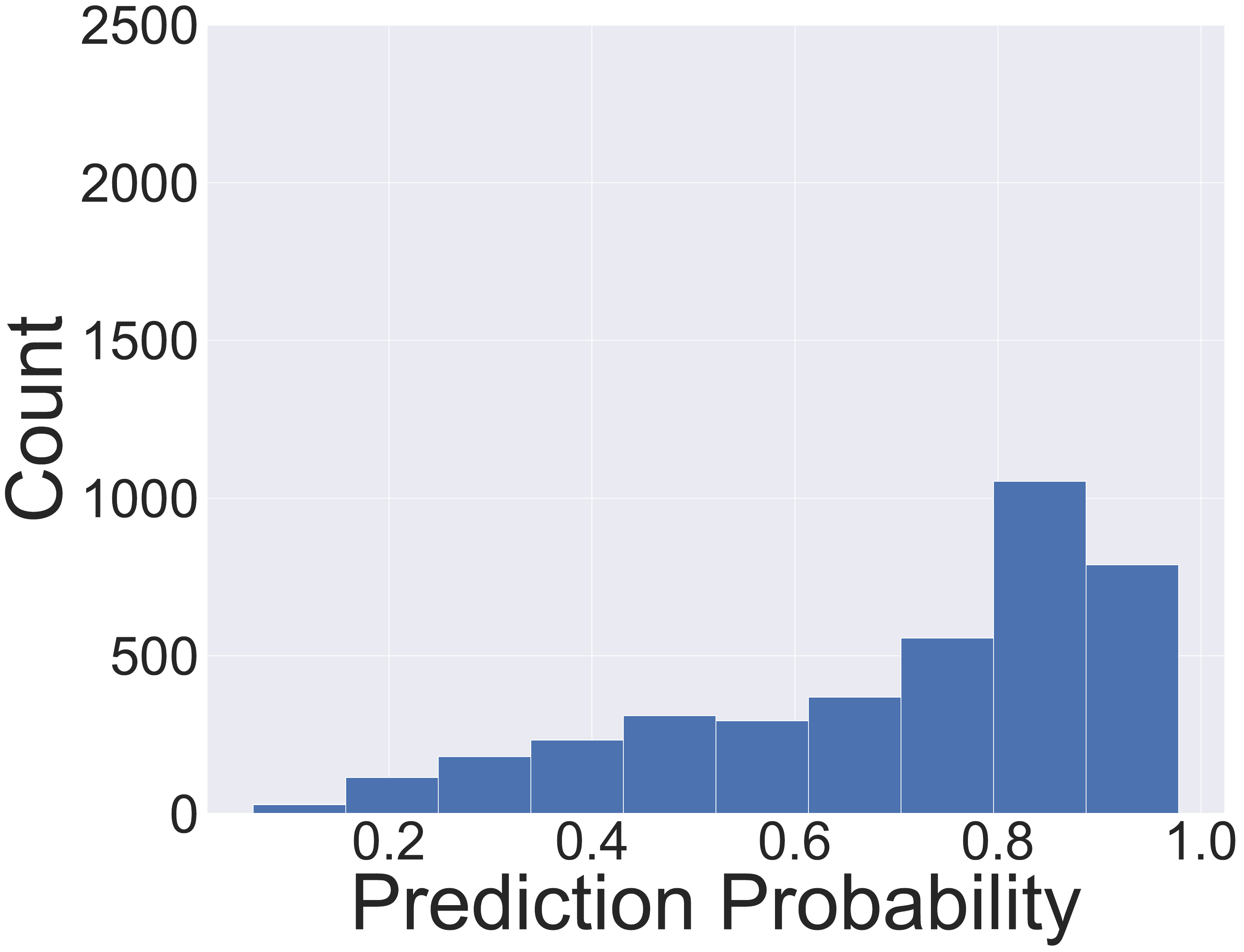}} &
        \subfloat[E-6 (F)]{\includegraphics[width=0.16\textwidth]{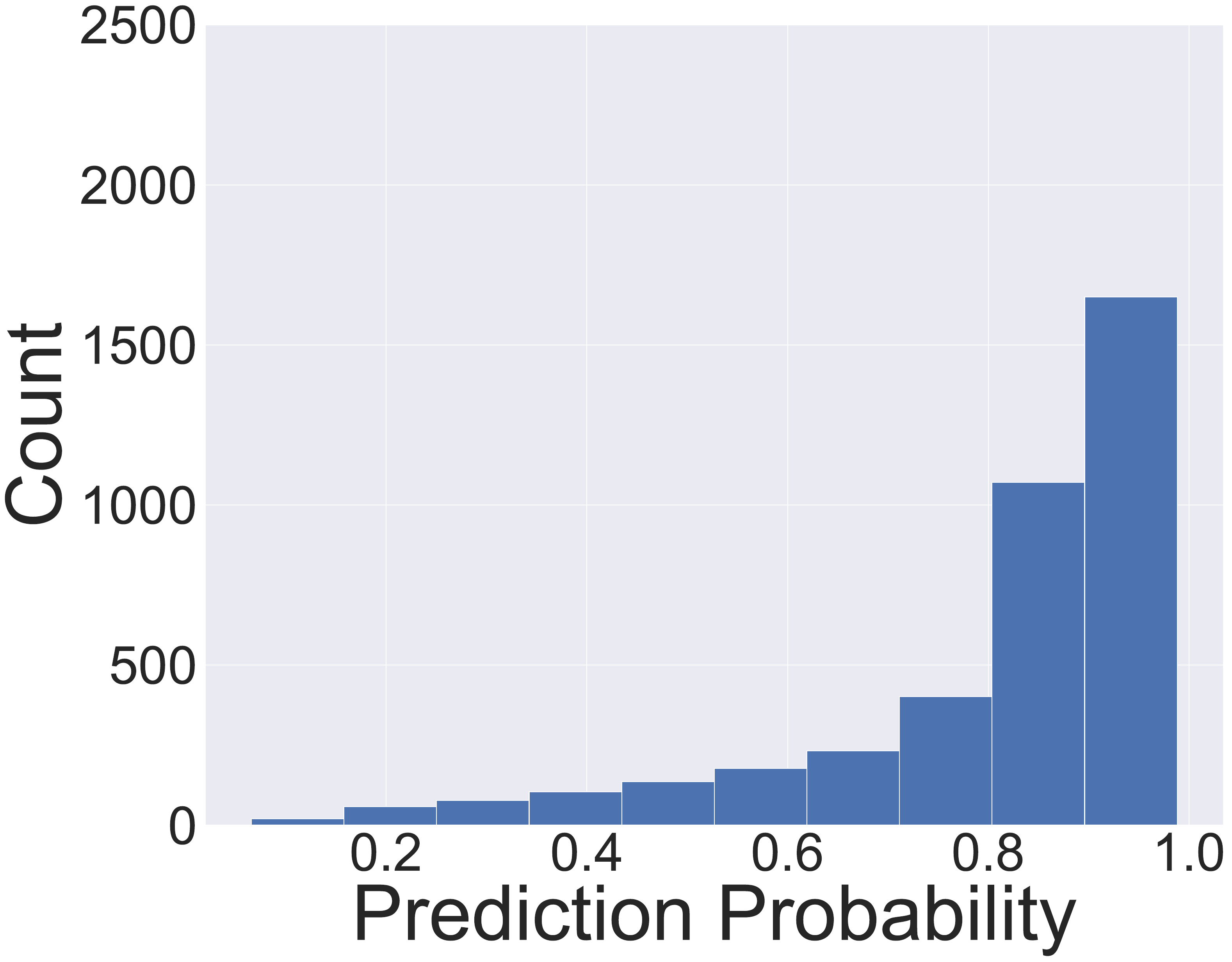}} &
        \subfloat[E-5 (I)]{\includegraphics[width=0.16\textwidth]{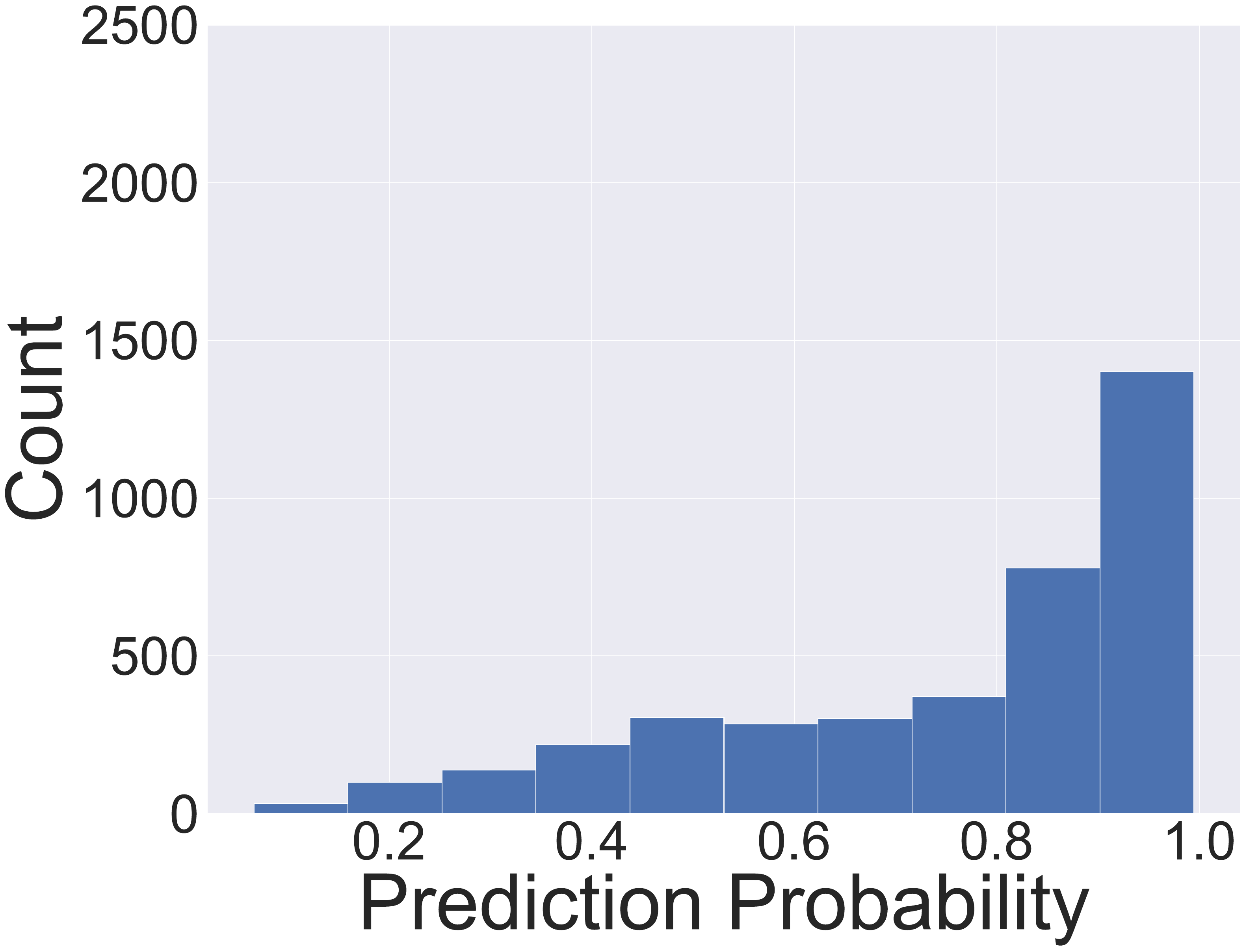}} &
        \subfloat[E-5 (F)]{\includegraphics[width=0.16\textwidth]{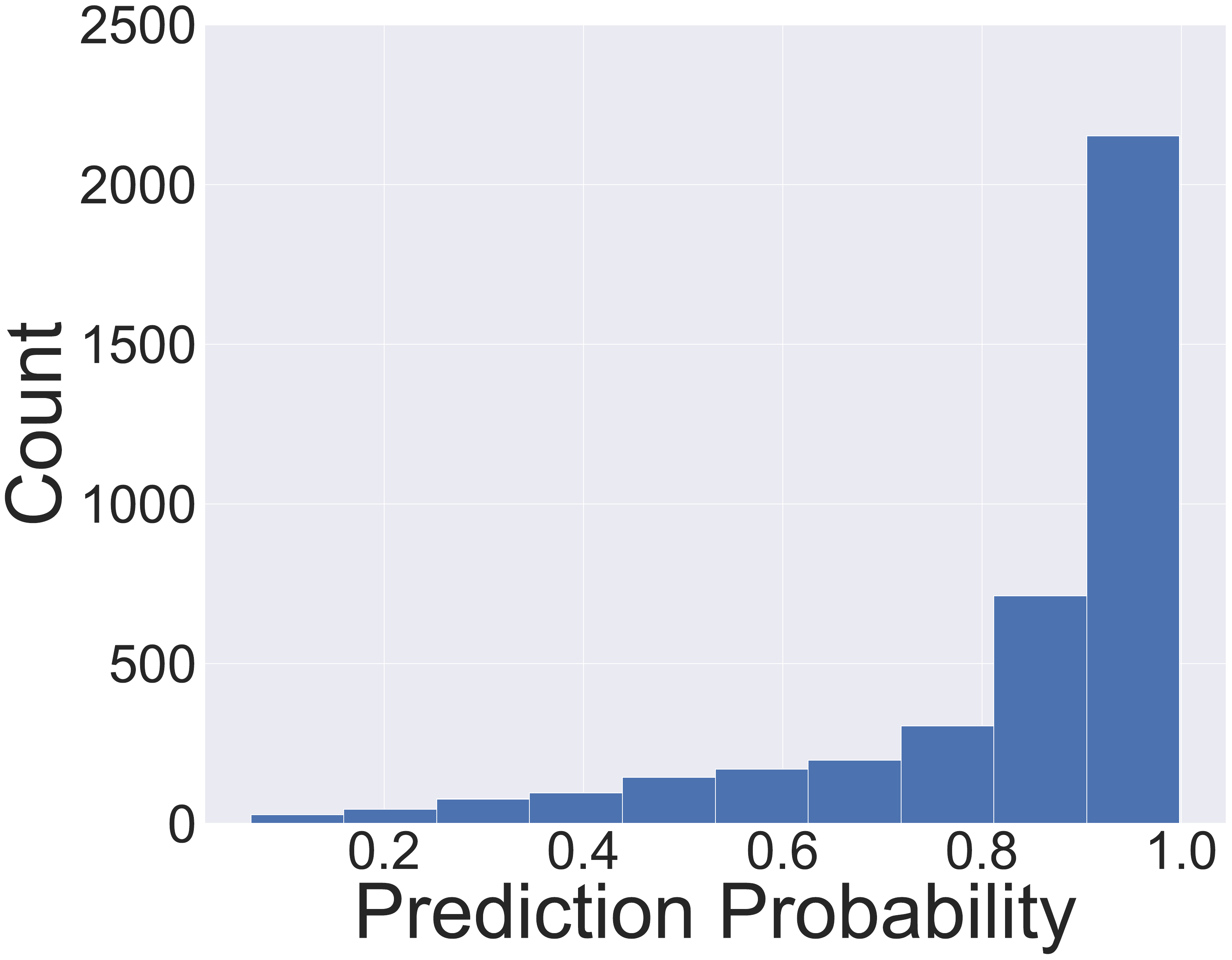}} \\
        \subfloat[ViT (I)]{\includegraphics[width=0.16\textwidth]{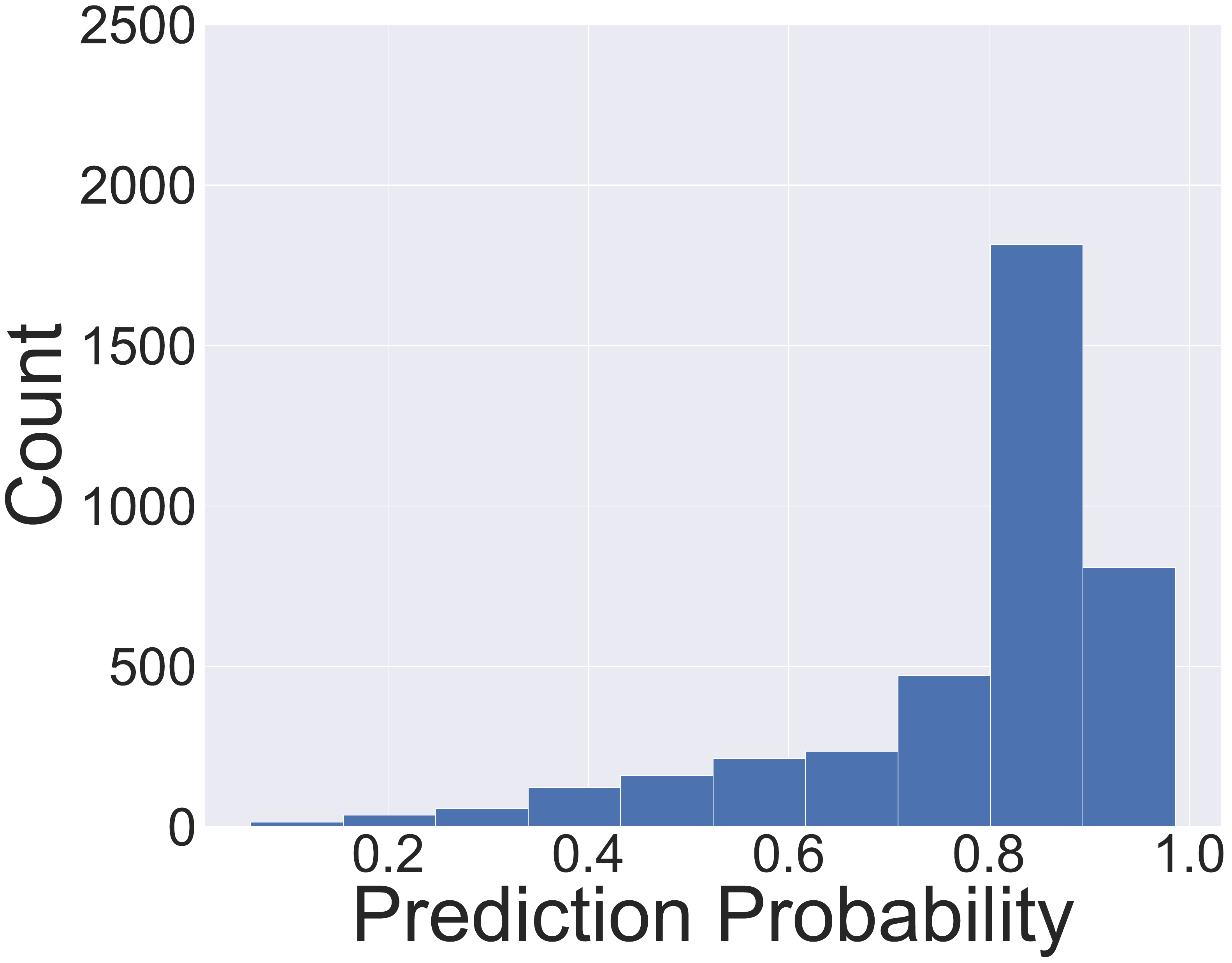}} &
        \subfloat[ViT (F)]{\includegraphics[width=0.16\textwidth]{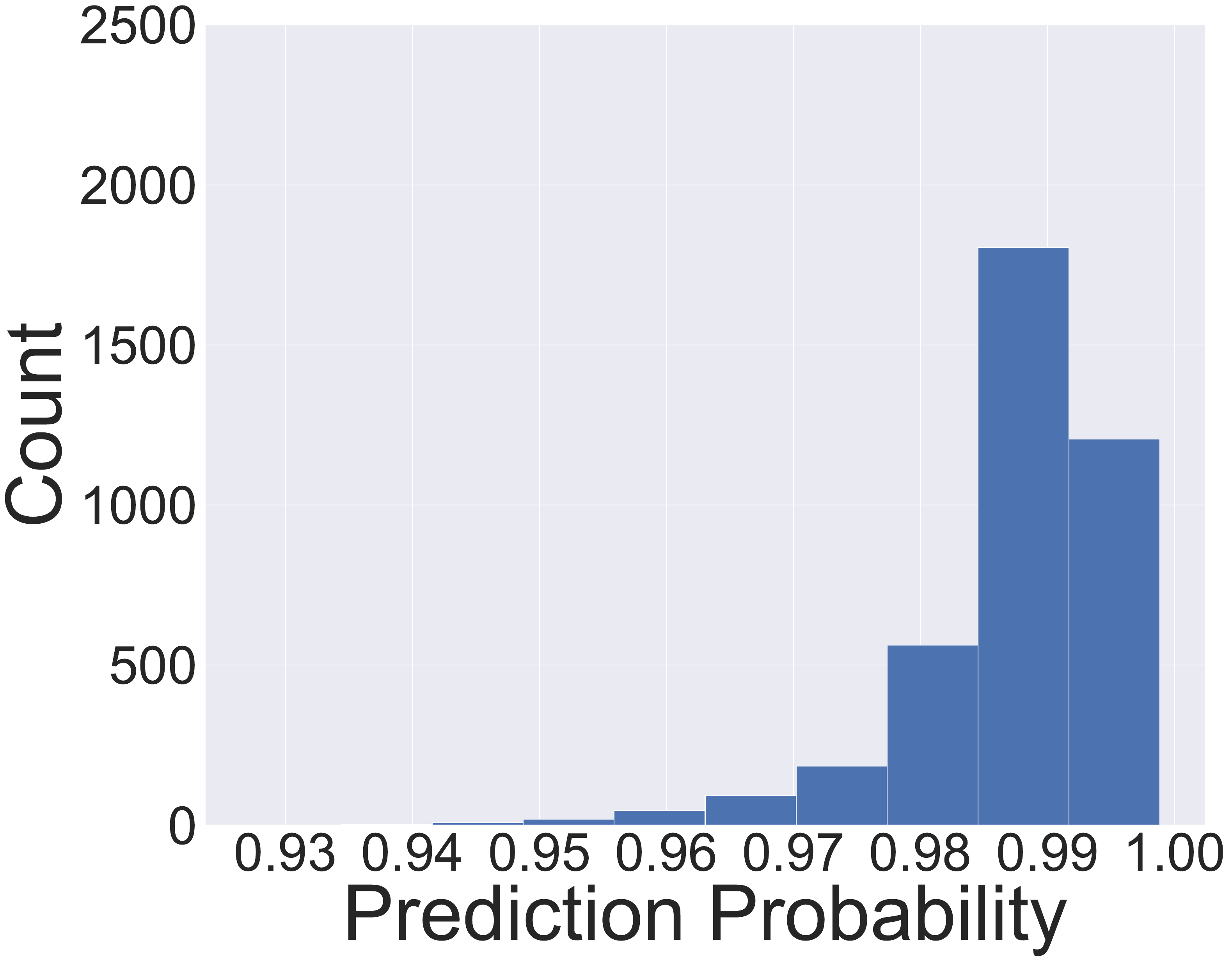}} &
        \subfloat[E-6 (I)]{\includegraphics[width=0.16\textwidth]{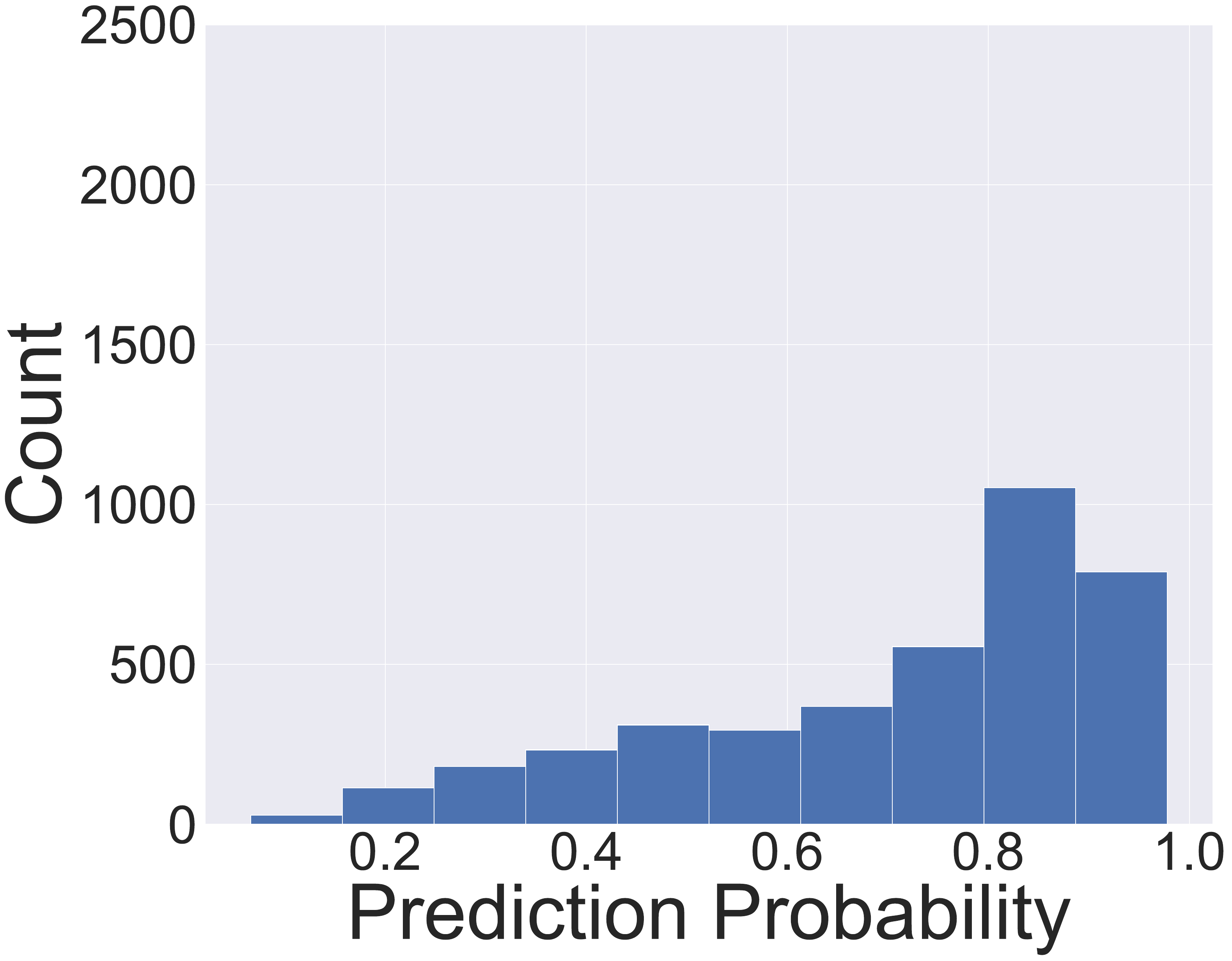}} &
        \subfloat[E-6 (F)]{\includegraphics[width=0.16\textwidth]{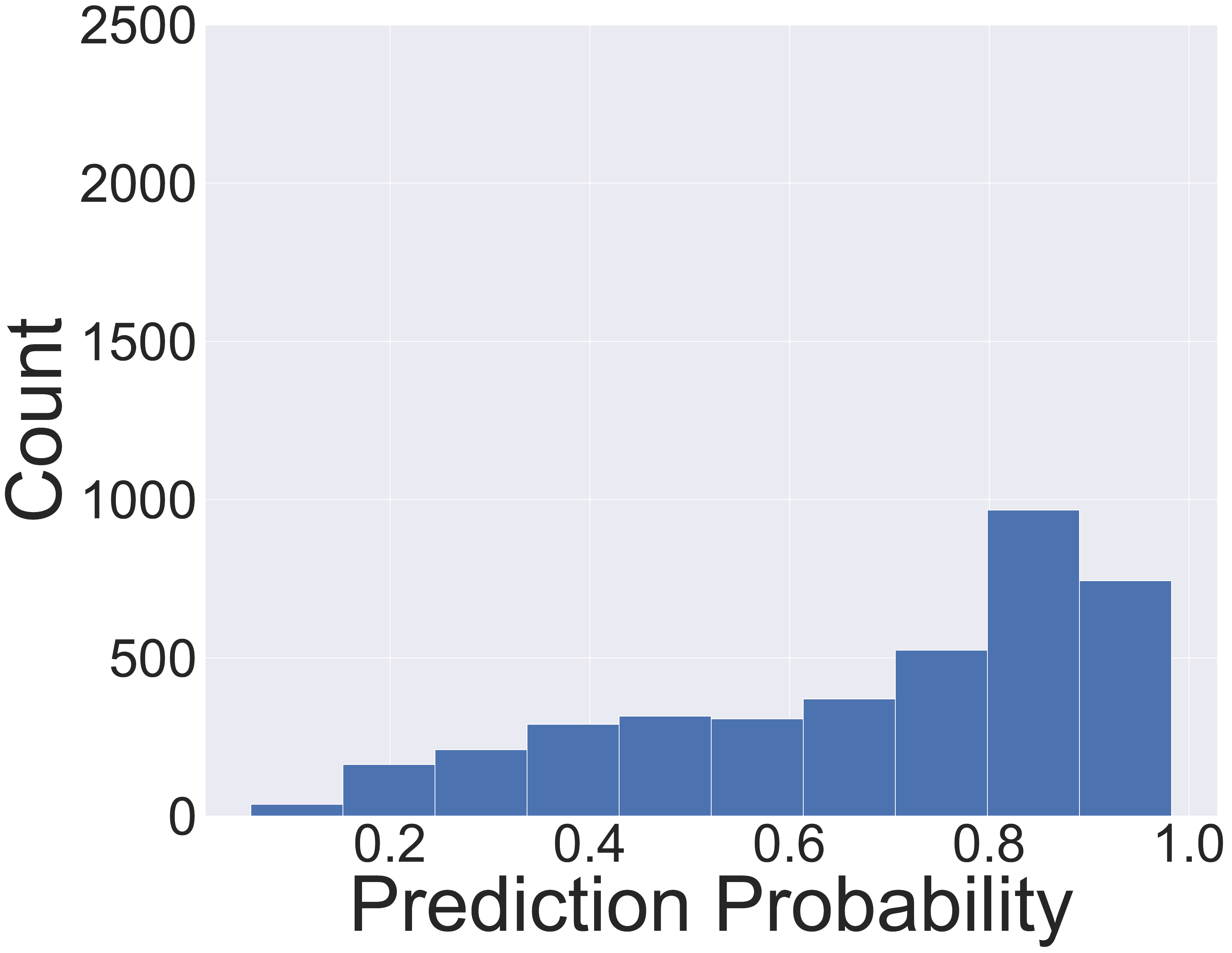}} &
        \subfloat[E-5 (I)]{\includegraphics[width=0.16\textwidth]{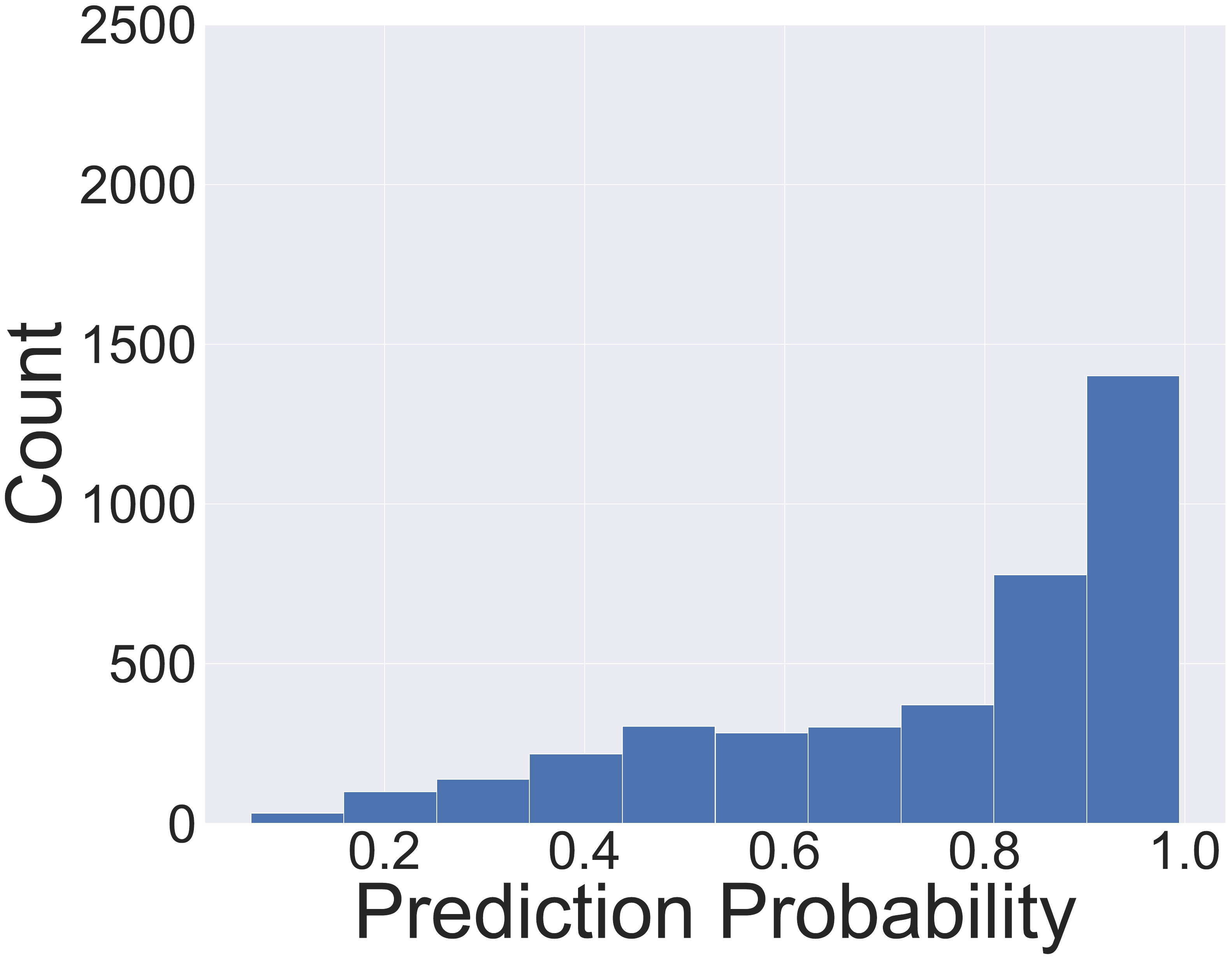}} &
        \subfloat[E-5 (F)]{\includegraphics[width=0.16\textwidth]{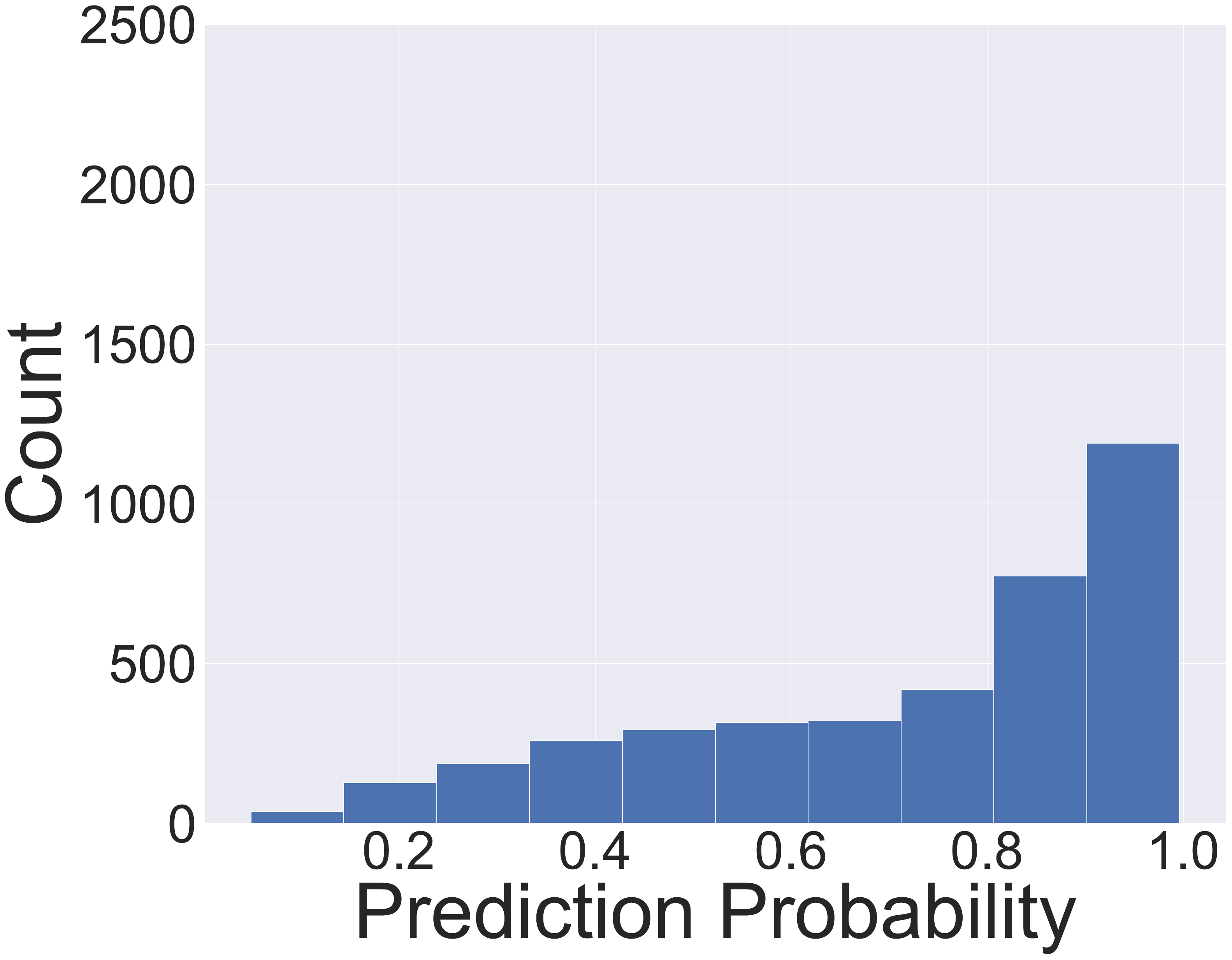}}
    \end{tabular}
    \caption{Histogram of class prediction probability distribution across different architectures, comparing image-only (I) and feature-only (F) inputs for the entire test dataset.}
    \label{fig:histogram}
\end{figure}

\begin{table}
\centering
\caption{The SS row examines model performance with image (I) and feature (F) inputs generated by the E-7 architecture and SS algorithm. It shows stable accuracy and F1 scores across the E-6 and E-5 architectures when using feature inputs. In contrast, the GC row, which uses the Grad-CAM algorithm for feature generation, demonstrates a drop in accuracy and F1 scores across the E-6, and E-5 architectures when features are used as input.}
\label{tab:e7}
\begin{tblr}{
  cell{2}{1} = {r=2}{},
  cell{4}{1} = {r=2}{},
  vlines,
  hline{1-2,4,6} = {-}{},
  hline{3,5} = {2-8}{},
}
Exp. & Metric     & E-7 (I)~ & E-7 (F)~ & E6 (I)  & E-6 (F) & E-5 (I) & E-5 (F) \\
SS        & Acc   & 78.4\%   & 74.09\%  & 75.90\% & 73.61\% & 77.27\% & 73.76\% \\
          & F1 & 0.87     & 0.84     & 0.85    & 0.84    & 0.86    & 0.84    \\
GM  & Accuracy    & 78.47\%  & 58.87\%  & 75.90\% & 55.74\% & 77.27\%    & 57.24\% \\
          &F1 & 0.87     & 0.72     & 0.85    & 0.70    & 0.86    & 0.71    
\end{tblr}
\end{table}
\begin{table}
\centering
\caption{The SS row evaluates model performance using image (I) and feature (F) inputs, generated through the ViT architecture and the SS algorithm. There's a slight decrease in accuracy and F1 score across architectures (E-6, E-5) with feature inputs. Similarly, the GC row, utilizing the Grad-CAM algorithm for feature generation, shows a comparable drop in accuracy and F1 scores across the E-6 and E-5 architectures when using feature inputs.}
\label{tab:vit}
\begin{tblr}{
  cell{2}{1} = {r=2}{},
  cell{4}{1} = {r=2}{},
  vlines,
  hline{1-2,4,6} = {-}{},
  hline{3,5} = {2-8}{},
}
Exp. & Metric      & ViT (I)~ & ViT (F)~ & E6 (I)  & E-6 (F) & E-5 (I) & E-5 (F) \\
SS        & Acc    & 89.92\%  & 88.83\%  & 75.90\% & 71.90\% & 77.27\% & 73.10\% \\
          & F1 & 0.94     & 0.94     & 0.85    & 0.83    & 0.86    & 0.83    \\
GM  & Ac    & 89.92\%  & 58.56\%  & 75.90\% & 44.21\% & 77.27\% & 43.47\% \\
          & F1 & 0.87     & 0.73     & 0.85    & 0.60    & 0.86    & 0.58    
\end{tblr}
\end{table}
Our experimental results (Tables \ref{tab:e7} and \ref{tab:vit}) indicate that features generated by a neural architecture can be detected by other architectures trained on the same data. This implies that feature attribution maps encapsulate sufficient data distribution information. Consequently, feature maps created using attribution maps on one architecture can be recognized by another architecture, provided that both are trained on the same data. As depicted in Fig. \ref{fig:histogram}, when we feed only features to the model, the class probability increases (Fig. \ref{fig:histogram} (b), (d), and (f)), particularly when using similar architectures for feature generation and evaluation. When employing different types of architectures (e.g., Transformer for generating feature maps and CNN for evaluating them), there is a slight drop in accuracy (Fig. \ref{fig:histogram} (j) and (l)), but the performance remains consistent.

Accuracy decreases when features are extracted with Grad-CAM saliency maps, suggesting these maps might not capture crucial information on the data distribution. However, when examining row GC in Tables \ref{tab:e7} and \ref{tab:vit}, it's observed that accuracy remains consistent across various architectural configurations when features are generated using Grad-CAM. This suggests that different architectures have harmony in detecting certain features from data.%
\section{Conclusion}
The experiment validates our hypothesis that various architectures acquire shared features from a common data distribution. We noticed a notable rise in class prediction probabilities when utilizing selected features as inputs, particularly when employing similar neural architecture building blocks such as Convolution. Additionally, the consistency of predictions on future attribution maps across architectures demonstrates that different architectures are not randomly learning features from the data, thereby enhancing the reliability of the models. These findings underscore the potential to generalize features and emphasize the need for additional research to harmonize feature attribution maps, expanding their applicability in various domains.

\subsubsection{Acknowledgements}
This work was partially funded by the German Federal Ministry of Education and Research (BMBF)
under grant number 16SV8639 (Ophthalmo-AI) and 2520DAT0P2 (XAINES) and German Federal Ministry
of Health (BMG) under grant number 2520DAT0P2 (pAItient) and supported by the Lower Saxony
Ministry of Science and Culture and the Endowed Chair of Applied Artificial Intelligence (AAI)
of the University
of Oldenburg.
\bibliographystyle{splncs04nat}
\bibliography{references}
\end{document}